\documentclass[letterpaper, 10 pt, conference]{ieeeconf}  % Comment this line out if you need a4paper

\usepackage{graphics}
\usepackage{amsmath} 
\usepackage{amssymb}
\usepackage{cite}
\usepackage{tikz}
\usepackage{amsmath}
\usepackage{tabularx}
\usepackage{array}
\usepackage{booktabs}
\usepackage{bbm}
\usepackage{adjustbox}
\usepackage{nicematrix}
\usepackage{caption}
\usepackage{gensymb}
\usepackage{svg}
\usepackage{flushend}
\usepackage{pifont}% http://ctan.org/pkg/pifont

\usepackage[ruled]{algorithm2e} % For algorithms

\SetKwInput{KwIn}{Input}
\SetKwInput{KwOut}{Output}
\SetKwFor{ForEach}{for each}{do}{end}

\usepackage{tablefootnote}

\usepackage{hyperref}

\IEEEoverridecommandlockouts                              % This command is only needed if 
\title{Understanding Fire Through \\Thermal Radiation Fields for Mobile Robots
}

\author{Anton R. Wagner$^{1}$,  Madhan B. Rao$^{2}$, Xuesu Xiao$^{2}$, S\"oren Pirk$^{1}$% <-this % stops a space
\thanks{$^{1}$Department of Computer Science, Kiel University, Germany        
        {\tt\small \{awa, sp\}@informatik.uni-kiel.de}}%
\thanks{$^{2}$ Department of Computer Science, George Mason University, USA
        {\tt\small \{mbalajir, xiao\}@gmu.edu}}%
}

\begin{document}

\maketitle
%%%%%%%%%%%%%%%%%%%%%%%%%%%%%%%%%%%%%%%%%%%%%%%%%%%%%%%%%%%%%%%%%%%%%%%%%%%%%%%%

\begin{abstract}

Safely moving through environments affected by fire is a critical capability for autonomous mobile robots deployed in disaster response. In this work, we present a novel approach for mobile robots to understand fire through building real-time thermal radiation fields. We register depth and thermal images to obtain a 3D point cloud annotated with temperature values. From these data, we identify fires and use the Stefan–Boltzmann law to approximate the thermal radiation in empty spaces. This enables the construction of a continuous thermal radiation field over the environment. We show that this representation can be used for robot navigation, where we embed thermal constraints into the cost map to compute collision-free and thermally safe paths. We validate our approach on a Boston Dynamics Spot robot in controlled experimental settings. Our experiments demonstrate the robot’s ability to avoid hazardous regions while still reaching navigation goals. Our approach paves the way toward mobile robots that can be autonomously deployed in fire-affected environments, with potential applications in search-and-rescue, firefighting, and hazardous material response.

\end{abstract}
%%%%%%%%%%%%%%%%%%%%%%%%%%%%%%%%%%%%%%%%%%%%%%%%%%%%%%%%%%%%%%%%%%%%%%%%%%%%%%%%
\section{INTRODUCTION}

Deploying robots in hazardous environments offers a safer alternative to direct human intervention, especially in disaster scenarios where conditions are unpredictable and dangerous. Firefighters and first responders face severe physical dangers when entering these environments, making autonomous robots a promising tool for scouting, victim search, situational assessment, and fire mitigation. However, fire-affected environments pose unique challenges due to the combined risks of smoke, heat, structural instability, and limited visibility. Elevated temperatures caused by fire can damage equipment, impair sensing, and render conventional navigation strategies insufficient, which primarily rely on geometric maps and obstacle avoidance, without explicitly accounting for fires. Safe operation in the presence of fire therefore requires mobile robots that not only respond to obstacles in their environment, but also understand fires and are aware of high temperatures. 

\begin{figure}[t]
    \centering
    \includegraphics[width=1.0\columnwidth]{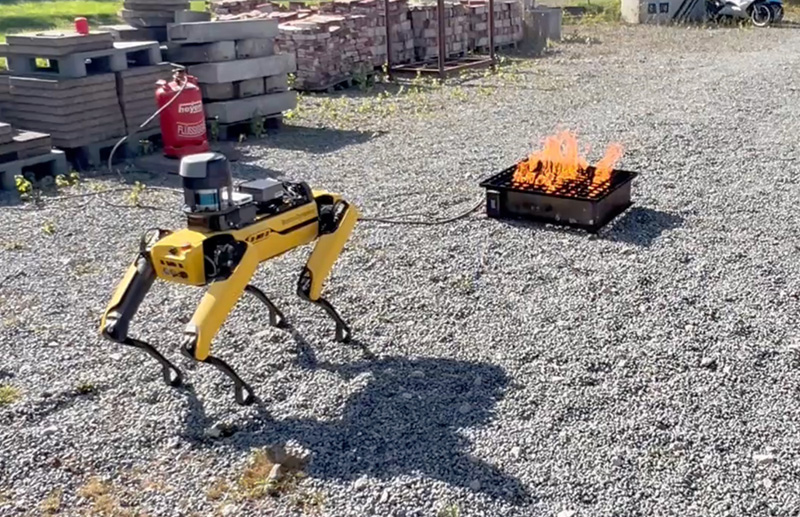}
    \caption{Experimental setup with the Boston Dynamics Spot robot approaching a controlled fire generated by a propane-based fire training device. The robot is equipped with depth and thermal sensors to perceive the environment and construct a thermal radiation field for fire-aware navigation.}
    \vspace{-5mm}
    \label{fig:setup}
\end{figure}

\begin{figure*}[t]
    \centering
    \includegraphics[width=1.0\textwidth]{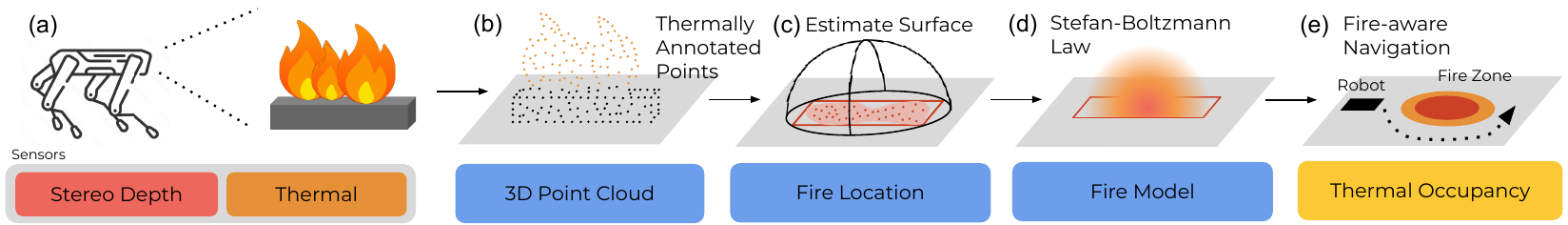}
    \caption{Overview of our framework: we use stereo depth and thermal sensor data to capture a fire (a) and to compute a thermally annotated point cloud (b). We use the points with the highest temperatures and project them into a 2D grid (c) and use the Stefan-Boltzmann law to estimate the heat decay of the fire (d). Finally, we integrate the computed thermal occupancy map with a common spatial occupancy map for fire-aware robot navigation (e).}
    \vspace{-5mm}
    \label{fig:overview}
\end{figure*}

To move around fire, it is desirable to measure temperatures between the robot and the fire source. However, measuring temperature in free space is challenging, as thermal cameras do not directly sense the air temperature or the heat a robot would encounter along a path. Instead, thermal cameras measure long-wave infrared radiation emitted from visible surfaces. As a result, the thermal image provides only surface temperatures of objects in the camera’s field of view, while the space between the robot and a heat source remains unobserved. This limitation makes thermal cameras insufficient on their own for navigation in fire-affected environments, since the most critical hazard -- the elevated temperatures between a fire source and the robot -- is what they cannot capture directly. 

To address this gap, we present an approach that constructs a thermal radiation field based on thermal perception  and a physical model that extrapolates from surface measurements to approximate the heat distribution throughout the environment to facilitate mobile robots' fire understanding. 
% In this work, we present a fire-aware navigation framework that uses thermal perception to compute temperature fields for the robot’s planning process. 
To be specific, we first register depth and thermal camera images to generate a 3D point cloud annotated with temperature values. High-temperature regions are then localized by fitting a spherical model to the thermal points, which provides a representation of the location of the fire source. Using the Stefan–Boltzmann law, we then model the radiative heat decay to construct a continuous thermal radiation field over the environment. This field is combined with geometric obstacles to form a thermally informed cost map, which we integrate into an A* planner for autonomous navigation. We validate our method on the Boston Dynamics Spot robot, demonstrating that the system can plan collision-free and thermally safe trajectories in fire-affected scenarios.

For our experiments (as shown in Fig.\ref{fig:setup}), we employ a fire extinguishing training device that generates controlled flames using a propane gas tank. This setup provides a realistic yet safe fire source with consistent thermal characteristics, allowing us to evaluate both perception and navigation under conditions resembling an actual fire scene. The Boston Dynamics Spot robot is tasked with walking toward the fire site, during which its sensors captured thermal and depth data to build the thermal radiation field used for planning.

In summary, our contributions are: (1) we register thermal and depth sensor images on a legged robot to produce 3D point clouds annotated with temperature measurements; (2) we detect and localize a fire by clustering high-temperature 3D points; and (3) we construct a continuous free-space radiative heat-flux field using a Stefan–Boltzmann-based power estimate for A*-based planning.

\section{RELATED WORK}
 We review related work in robotics and fire modeling to identify the research gap and motivate our contribution. 
 
\subsection{Robotics in Firefighting}
Operating robots in hazardous environments has a long history. 
A number of early systems targeted fire response tasks such as evacuation guide robots \cite{Kim2009}, human–robot swarm interaction studies for firefighting \cite{Naghsh2008}, and more recently, UAV swarm-based firefighting concepts \cite{Tzoumas2024,McConville2024}. Legged robots have also been investigated for firefighting, such as quadruped platforms carrying fire suppression systems \cite{Baird2024}. These works highlight the potential of deploying robotic systems in fire scenarios, but they typically do not endow robots with an explicit notion of thermal fields.

\subsection{Thermal Sensing for Mapping and Navigation}
Thermal sensing has been widely used in mapping and inspection tasks. Cross-modal calibration between thermal and 3D sensors is an important enabler~\cite{datar2025m2p2}. Fu et al. \cite{Fu2021} proposed a target-less method for extrinsic calibration of stereo, thermal, and LiDAR sensors directly in the field, addressing a key bottleneck for practical deployment. 
A series of works demonstrated 3D thermography from depth–thermal fusion \cite{Borrmann2014,Vidas2013} and radiometric LiDAR–thermal mapping~\cite{DePazzi2022}. These methods show that temperature measurements can be projected into a 3D reference frame, creating dense thermal reconstructions of 3D scenes.

For navigation in degraded environments~\cite{datar2025m2p2}, thermal SLAM has become an active research direction. Direct thermal-infrared SLAM with sparse depth cues has been demonstrated \cite{Shin2019} and infrared–LiDAR edge-based SLAM has been proposed to improve robustness~\cite{Chen2021}. FirebotSLAM explicitly addressed smoke-filled environments by exploiting thermal imaging for localization \cite{vanManen2023}. Beyond traditional SLAM, Saputra et al. \cite{Saputra2021} introduced a graph-based thermal–inertial SLAM framework that integrates probabilistic neural networks to improve performance in low-visibility conditions. Together, these approaches show that thermal sensing can support robust localization when conventional cameras fail.

Firefighting scenarios pose unique challenges: smoke, thermal shimmer, and high-intensity flames degrade conventional sensors such as RGB and LiDAR, motivating the usage of thermal sensing for mapping and navigation \cite{Starr2014,Wellhausen2020SafeRN}. However, thermal representations from most existing thermal reconstruction and localization approaches only apply to objects, not free spaces, which is insufficient for navigation in active fires. 

\subsection{High-Fidelity Environmental Modelling}
Research in visual computing has recently targeted higher-fidelity physical reconstructions. Chu et al. used physics-informed neural fields to model smoke with sparse data \cite{Chu2022}, while Jin et al. \cite{Jin2025} introduced ActiveGS for scene reconstruction with Gaussian splatting. These efforts improve understanding of fire and smoke dynamics but are computationally heavy and not designed for real-time navigation.

\subsection{Research Gap and Our Contribution}
The primary limitation of these works is that they model the surface temperature of objects. This information is insufficient for safe robot navigation, as the radiant heat from the fire can propagate through the free space which is not captured by surface-centric thermal maps. Our work addresses this gap by using the surface temperatures to model the continuous thermal radiation field in surrounding space. Unlike prior works, our contribution is navigation-oriented: we localize the fire source, construct a continuous and physically grounded thermal radiation field using the Stefan–Boltzmann law, and inject this field directly into the navigation cost function of a mobile robot. To our knowledge, no previous approach explicitly models thermal radiation fields from thermal–depth data for real-time path planning in active fire scenarios.
\section{METHOD}

In this section, we describe our thermal radiation field representation for fire-aware navigation, which combines thermal–depth sensor registration, fire localization through geometric fitting, thermal radiation field modeling based on the Stefan–Boltzmann law, and integration of this field into a planning pipeline executed on a Boston Dynamics Spot robot. An overview of our framework is shown in Fig.~\ref{fig:overview}, the used symbols and their units are shown in Tab.~\ref{tab:symbols}.

\subsection{Thermal-Depth Fusion}

To enable fire localization and thermal radiation field estimation, we first register the thermal camera mounted on the front top of the Spot robot with its onboard depth sensing system. We use the stereo camera of the Spot for depth estimation as this allows us to get depth values for flames. While the thermal camera produces a 2D image of infrared intensities, the stereo cameras output a 3D point cloud in the reference frame of the robot. We formulate this registration as an extrinsic calibration problem, where the goal is to find the rigid-body transformation (rotation and translation) that maps rays from the thermal camera frame into the coordinate system of the depth camera. A prerequisite for this step is the intrinsic calibration of the thermal camera, where we estimate its camera matrix $K$ containing the focal lengths and principal point, along with distortion parameters. To obtain these values, we employ a calibration target that produces strong thermal contrast (i.e., a heated checkerboard), allowing robust corner detection in thermal imagery and consequently the robust registration of RGB, depth, and thermal frames (Fig. \ref{fig:registration_overlay}). 

With the computed $K$ the thermal image can be projected onto the depth values. 
For intrinsic calibration of the thermal camera, we used a heated checkerboard pattern, while extrinsic calibration between the stereo pair and the thermal camera was performed using an ArUco board (Fig.~\ref{fig:registration}). This choice was motivated by the thermal camera’s low resolution and narrow field of view, which made it impractical to capture the full checkerboard at close range. In contrast, ArUco markers only require a subset of visible markers in each modality, making them more robust for this setup. We used {Calibmar toolkit~\cite{seegraeber2024} to compute the extrinsic parameters from a set of matching RGB and thermal images at different angles and positions in both cameras' frames. After registering the thermal and stereo-camera data, we obtain a 3D point cloud annotated with temperature values. This fused representation allows us to jointly reason about geometry and heat in the robot’s environment, which enables us to localize fire and in turn to construct a thermal radiation field for navigation. In Fig. \ref{fig:fire_on} we show the measured temperature values from the perspective of the robot before and after turning on a fire. 

\subsection{Thermal Radiation Fields} 
To identify the fire, we cluster all 3D points with temperatures above $100\degree C~(373.15 K)$ using DBSCAN with parameters $\epsilon=0.5$ and \text{min\_samples} = 2. The cluster with the largest number of points is selected as the fire source. From the two largest spatial extents of this cluster, we estimate the fire footprint by circumscribing the resulting rectangle and fit a hemisphere above it (Fig.\ref{fig:overview}b, c). This hemisphere provides a surface area $A$ that we use to estimate the radiative power of the fire with the Stefan–Boltzmann law (Fig.\ref{fig:overview}d). We assume a fixed fire temperature of
\begin{equation}
T_0 = 1,473.15 \mathrm{K} \quad (1,200^{\circ} \mathrm{C}),
\end{equation}
which is below the actual flame temperature of a stoichiometric propane fire ($1,980 \degree C$) to account for non-ideal combustion conditions. The thermal radiation emitted from the fire surface is
\begin{equation}
P = \sigma A T_0^4\gamma,
\label{eq:equation2}
\end{equation}
where $\sigma$ is the Stefan–Boltzmann constant and $\gamma$ a correction factor set to $\gamma=0.4$, that we empirically found to provide reasonable estimates for the total fire power. At distance $r$ from the fire center, the thermal radiation decays according to the inverse-square law: 
\begin{equation}
P(r) = \frac{P}{4 \pi r^2} \mathcal{X},
\end{equation}
where $\mathcal{X}$ is the fraction of power emitted as thermal radiation that we set to $\mathcal{X}=0.35$ according to Hamins et al.~\cite{hamins1999nist}. 
This thermal radiation distribution is sampled onto a 2D grid representing the environment (20 x 20m) at 10 cm resolution, yielding a 200×200 grid. We initialize the grid with occupancy values from LiDAR, with
\begin{equation}
\mathcal{O}[x,y] =
\begin{cases}
1 & \text{if obstacle height } > h_\text{robot}, \\
0 & \text{otherwise},
\end{cases}
\end{equation}
where $h_{robot}$ defines the height of the robot -- anything taller than the robot is defined to be occupied. For each grid cell, we evaluate $P(r)$ at the cell center. 

\begin{figure}[t]
    \centering
    \includegraphics[width=1.0\columnwidth]{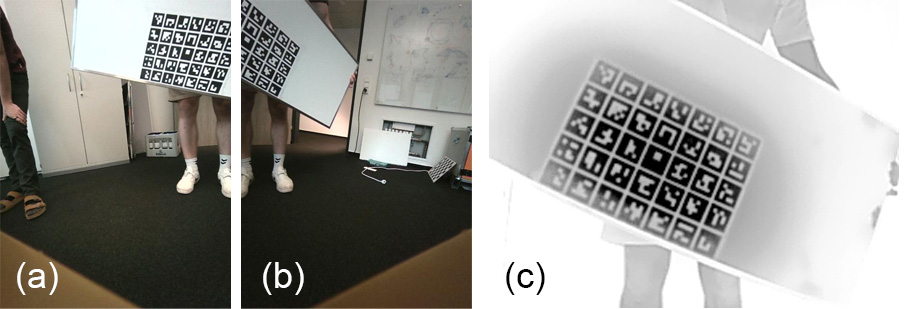}
    \caption{To register the right~(a) and left~(b) RGB frames of the Spot robot and of the thermal~(c) camera, we use the Calibmar toolkit~\cite{seegraeber2024}. We use a heated ArUco board to obtain a thermal frame with significant thermal differences between the white and black areas.}
    \vspace{-4mm}
    \label{fig:registration}
\end{figure}
\begin{figure}[t]
    \centering
    \includegraphics[width=1.0\columnwidth]{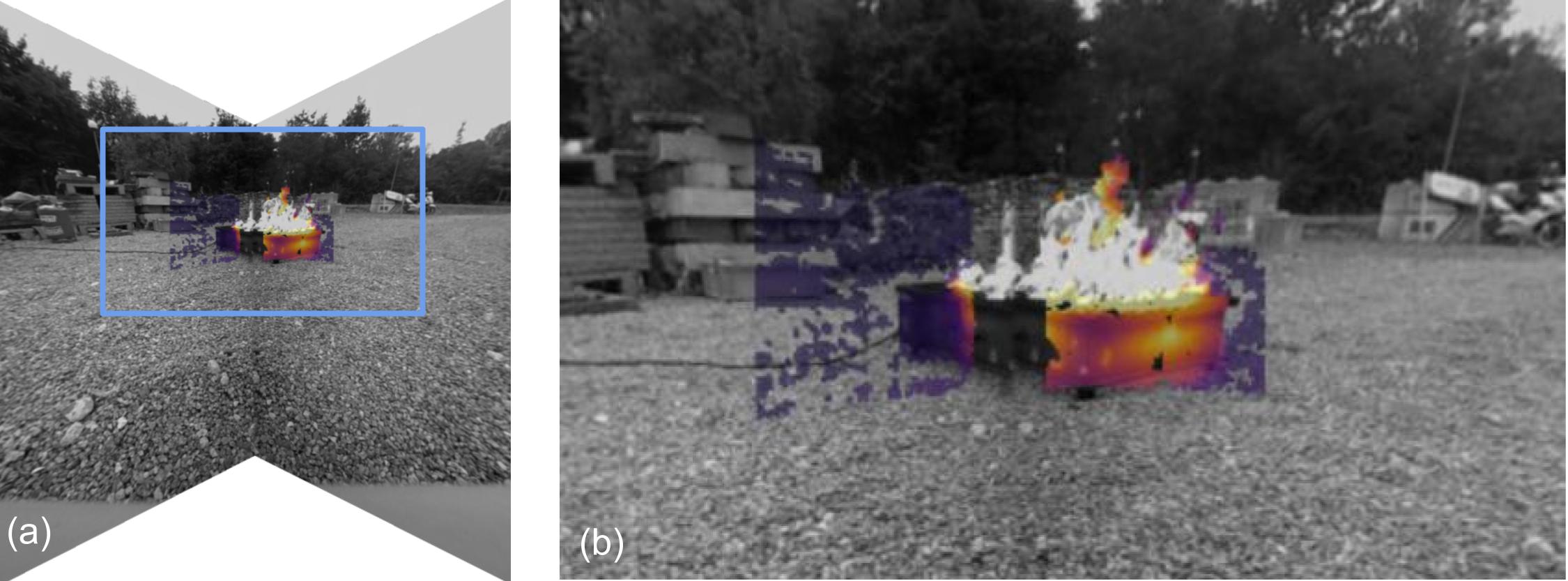}
    \caption{Registration result: we show the registered RGB frames of the Spot with a thermal overlay (a) as well as a zoomed in view to show more details (b). The RGB frames have been converted to a grayscale image to show the color coded thermal overlay.}
    \vspace{-2mm}
    \label{fig:registration_overlay}
\end{figure}

\begin{figure}[t]
    \centering
    \includegraphics[width=1.0\columnwidth]{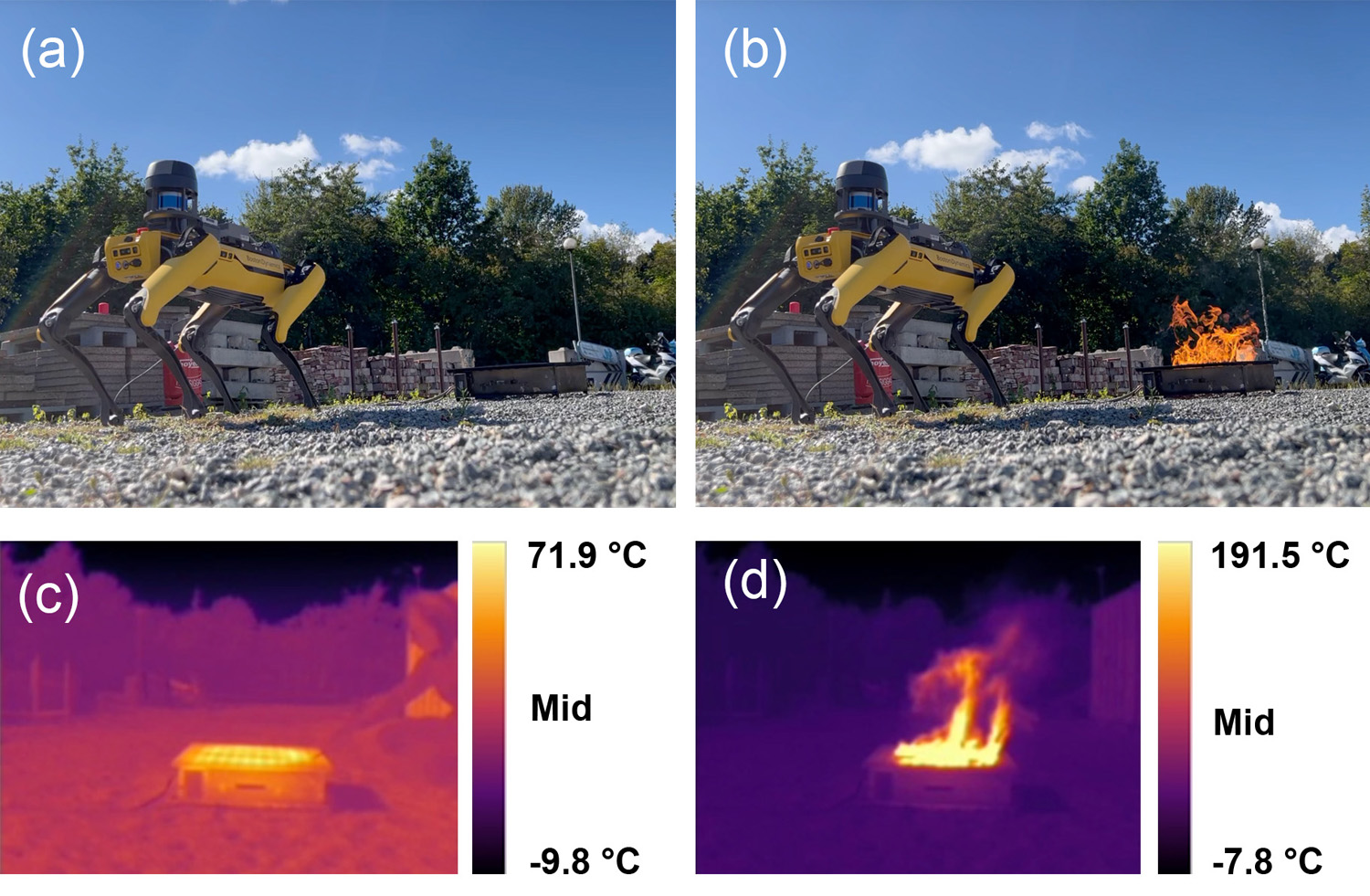}
    \caption{Experimental setup showing the Boston Dynamics Spot robot and the fire source turned off (a) and turned on (b). The thermal images taken from the perspective of the robot show the corresponding temperatures obtained with the thermal camera (c), (d).}
    \vspace{-4mm}
    \label{fig:fire_on}
\end{figure}

To account for occlusions, we perform ray tracing between the fire source and each grid cell using a Bresenham algorithm. A binary line-of-sight function $\mathrm{LoS}(\cdot) \in \{0,1\}$ is used to model visibility, where a value of 0 indicates occlusion and 1 indicates a clear line of sight. Radiation is stored in a thermal grid when a line of sight from the cell center to the fire exists:  
\begin{equation}
\mathcal{T}[x,y] = P(r).
\end{equation}
Additionally, we define a threshold at which we deem the space impassable due to thermal radiation:
\begin{equation}
 q_\text{danger} = \frac{2.5}{max(0.1, \phi)},
\end{equation}
where $\phi$ defines a scalar value to tune the desired navigation behavior. 
Finally, the occupancy grid is updated by blending geometric and thermal hazards:
\begin{equation}
\mathcal{O}[x,y] = \max\Big( \mathcal{O}[x,y], ; \min\Big(\frac{\mathcal{T}[x,y]}{q_\text{danger}}, 1.0\Big) \Big),
\end{equation}
In combination, this allows us to consider that tall obstacles block heat propagation. While thermal penalties increase linearly with increased thermal radiation, thermal radiation increases quadratically with proximity to the fire. 

\begin{figure*}[t]
    \centering
    \includegraphics[width=1.0\textwidth]{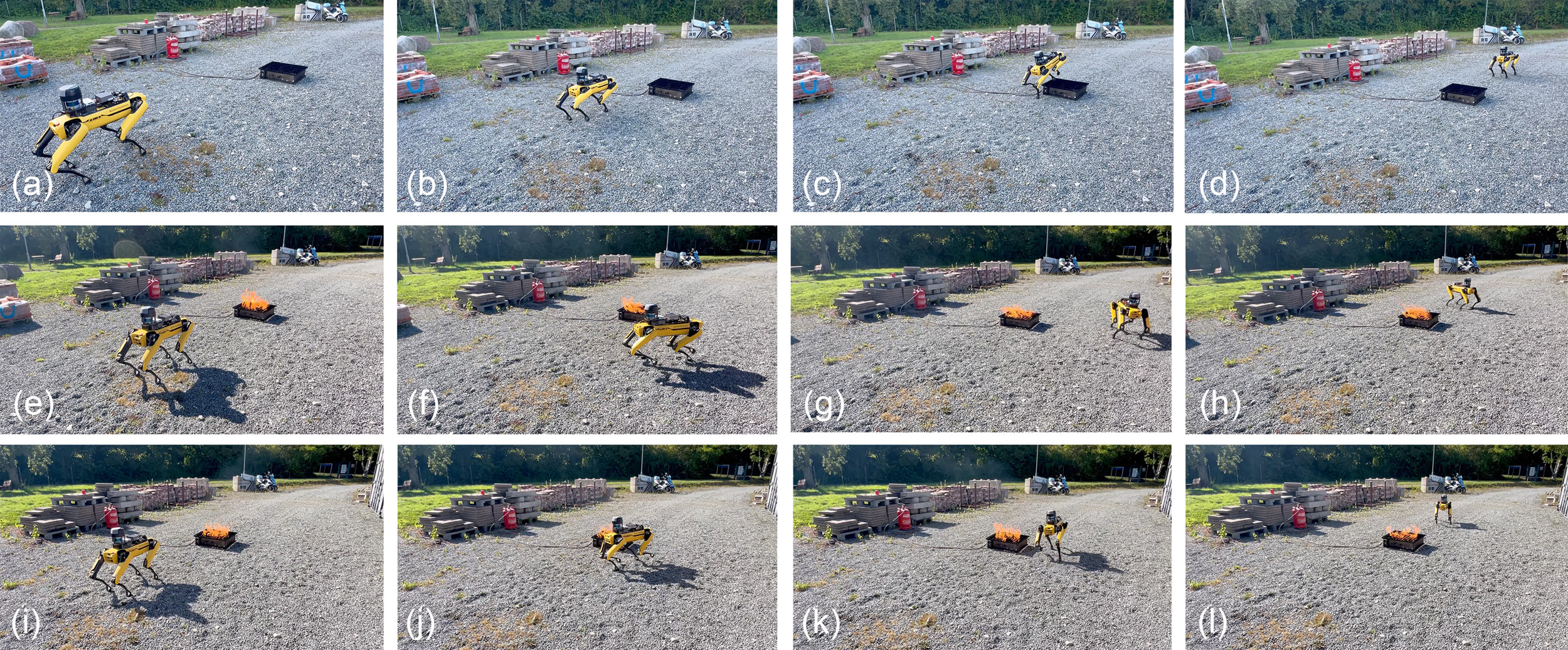}
    \caption{Fire-aware navigation results for the Boston Dynamics Spot robot. When no fire is present the robot walks over the fire training device (a)-(d). When the fire is turned on and a larger safety margin is activated, the robot plans and executes a trajectory further away from the fire (e)–(h). When a smaller safety margin is activated, the robot takes a path closer to the fire (i)–(l).}
    \vspace{-6mm}
    \label{fig:navigation}
\end{figure*}

\begin{algorithm}[t]
\small
\SetAlgoLined
\LinesNumbered
\DontPrintSemicolon
\SetAlgoVlined
\KwIn{LiDAR scans, robot positions, caution, optional fire location $\mathcal{L}$ and intensity $P$.}
\KwOut{2D occupancy grid for navigation.}

Initialize $grid\_size \gets map\_size\_m / resolution$\;
Initialize $\mathcal{O}[0:grid\_size,0:grid\_size] \gets 0$\;
Initialize $height\_grid[0:grid\_size,0:grid\_size] \gets -\infty$\;
Initialize $obstacle[0:grid\_size,0:grid\_size] \gets False$\;
Initialize $\mathcal{T}[0:grid\_size,0:grid\_size] \gets 0$\;
$q_\text{danger} \gets \frac{2.5}{max(0.1, \text{caution})}$\;

\tcp{Process LiDAR scans}
\ForEach{LiDAR scan}{
    \ForEach{point $(x,y,z)$ in scan}{
        Convert $(x,y)$ to grid indices $(grid_x, grid_y)$\;
        \If{indices inside grid}{
            $height\_grid[grid_y,grid_x] \gets \max(height\_grid[grid_y,grid_x], z)$\;
        }
    }
}

\tcp{Compute height-based occupancy}
\ForEach{grid cell $(i,j)$}{
    \If{$height\_grid[i,j] \neq -\infty$}{
        Collect heights 10m around robot\;
        $local\_ground \gets 20$th percentile of local heights\;
        \If{$height\_grid[i,j] - local\_ground > h_{robot}$}{
            $obstacle[i,j] \gets True$\;
            $\mathcal{O}[i,j] = 1.0$
        }
    }
}

\tcp{Compute radiative heating from fire}
\If{$\mathcal{L}$ and $P$ are passed in}{
    \ForEach{grid cell $(i,j)$}{
        Convert grid indices to world coordinates\;
        Compute world distance $d$ from $\mathcal{L}$\;
        \If{obstacle-free line between $\mathcal{L}$ and cell}{
            $\mathcal{T}[i,j] \gets P / (4 \pi d^2)$\;
        }
    }
}

\tcp{Generate final occupancy map}
\ForEach{grid cell $(i,j)$}{
        $\mathcal{O}[i,j] \gets \max(\mathcal{O}[i,j], \min(\mathcal{T}[i,j] / q_\text{danger}, 1.0))$\;
}
\KwRet $\mathcal{O}$\;

\caption{Occupancy map generation with LiDAR and thermal radiation fields.}
\label{alg:occupancy_map}
\end{algorithm}

\subsection{Robot Navigation}

The fire-aware navigation framework integrates the thermal radiation fields into the robot’s motion planning pipeline. The Boston Dynamics Spot robot provides proprioceptive state estimation and collision avoidance through its onboard autonomy stack, while our method augments this capability with a thermally informed cost map. Specifically, the environment is represented as a 2D occupancy grid derived from the registered depth data, onto which we overlay the continuous thermal radiation field computed from the fitted fire model. Each grid cell is assigned a navigation cost that combines geometric occupancy with thermal exposure, such that regions closer to the fire yield higher penalties.

Formally, the cost of traversing a grid cell is defined as

\begin{equation}
C[i,j] =
\begin{cases}
\infty & \text{if $\mathcal{O}[i,j]$ } = 1.0, \\
1.0 + \mathcal{O}[i,j]\beta & \text{otherwise},
\end{cases}
\end{equation}
where $\beta$ is a weighting factor that adjusts the relative importance of thermal safety versus path efficiency. The thermal cost is defined from the modeled thermal radiation field 
$T(r)$, such that higher values correspond to regions closer to the fire source. Other obstacles and areas with a thermal radiation $\mathcal{T}[x, y]>q_\text{danger}$ are impassable. 

To compute safe paths, we employ the A* search algorithm, which explores the cost map to find a trajectory from the robot’s current pose to the desired goal. In this formulation, the planner naturally trades off between geometric distance and thermal risk: paths that approach the fire source too closely are discouraged by high cost values, while longer detours around the fire remain feasible solutions. During execution, Spot follows the planned path using its built-in locomotion controller, ensuring stable traversal over uneven ground while respecting the thermal constraints.

In practice, this approach allows the robot to adjust its trajectory based on the specified safety margin. Larger safety margins increase the thermal penalties near the fire, resulting in paths that maintain greater distance from the heat source, while smaller margins yield more direct paths that pass closer to the fire. This flexibility demonstrates how incorporating the thermal radiation field into the cost map enables explicit control over the balance between efficiency and safety in fire-affected environments.

\subsection{Algorithmics}

Algorithm \ref{alg:occupancy_map} provides an overview of the implementation for constructing a fire-aware occupancy map by fusing geometric data from LiDAR with thermal radiation fields. We start by initializing a grid representation of the environment, in which each cell stores height, occupancy, and thermal values. LiDAR scans are then processed to record the maximum observed height in each cell, which is used to classify taller structures as occupied obstacles. To capture the effect of fire, the algorithm computes thermal radiation for all grid cells using the Stefan–Boltzmann model and the inverse-square law, while enforcing line-of-sight constraints to ensure that heat does not propagate through occlusions. Cells exposed to radiation are assigned intermediate costs, while those exceeding a danger threshold are considered hazardous and therefore impassible. In the final step, the algorithm blends geometric and thermal information by taking the maximum occupancy value per cell, thus integrating both physical obstacles and radiative hazards into a unified map. This fire-aware occupancy grid serves as the input to the planner, enabling the robot to compute paths that are both collision-free and thermally safe.

\section{EXPERIMENTS and VALIDATION}

To validate the proposed approach, we conducted a set of controlled experiments designed to test how our thermal radiation field representation can enable the robot to perceive thermal hazards, conduct fire-aware navigation, and adapt its motion planning. Additionally, we validated our thermal radiation fields with a calorimetric experiment. 

\begin{figure*}[t]
    \centering
    \includegraphics[width=1.0\textwidth]{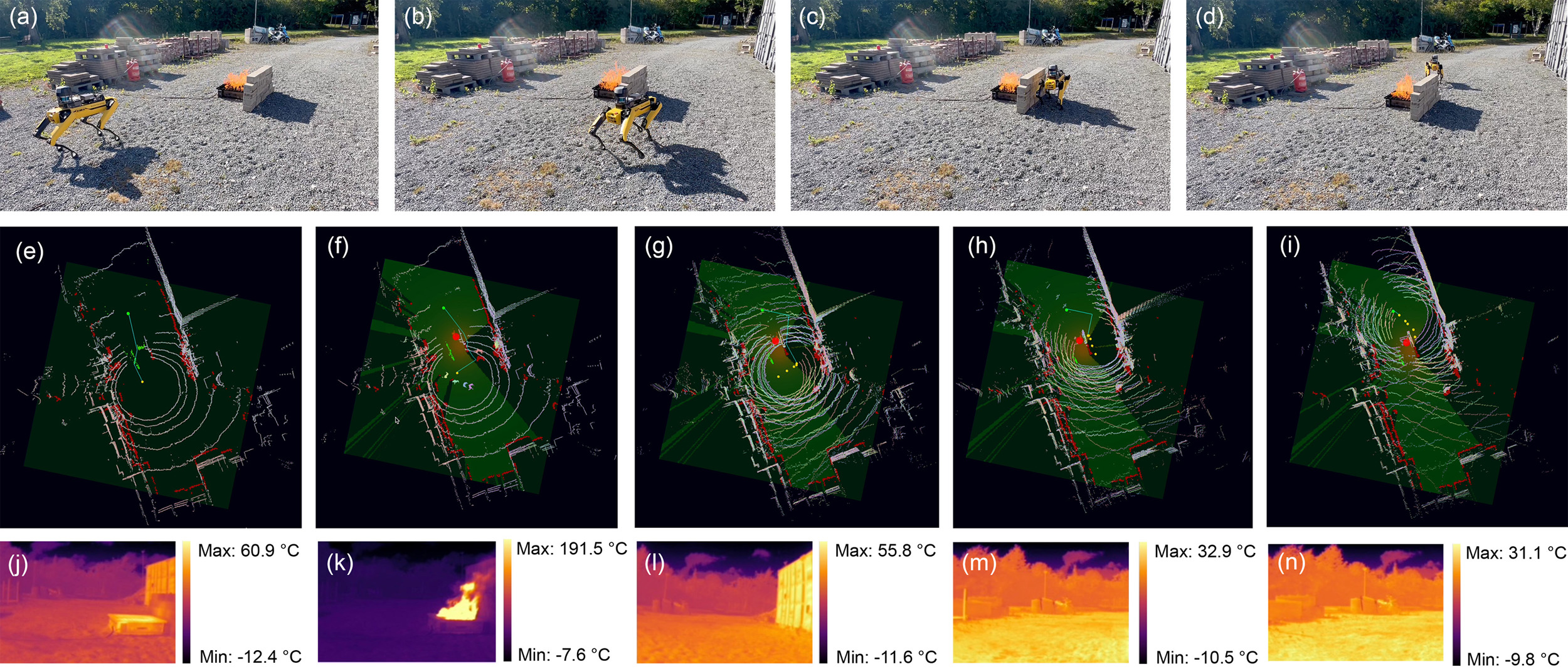}
    \vspace{-4mm}
    \caption{Thermal occlusion: If an obstacle blocks the fire source (e.g., a wall), our approach dynamically adjusts and computes a new path around the fire (a)-(d). Here we show the planned path (green) before the fire source was turned on (e) and the newly computed path also considering the safe space adjacent to the wall (f) that the robot then traverses (g)-(d). The captured thermal images show the measured temperatures along the path (j)-(n).}
    \vspace{-5mm}
    \label{fig:wall}
\end{figure*}

\subsection{Experiments}
We designed a series of experiments to evaluate our fire-aware navigation framework on the Boston Dynamics Spot robot in a controlled setting with a single fire source (Fig. \ref{fig:setup}). The goal of these experiments is to assess the robot’s ability to account for thermal hazards for path planning and to adapt its trajectories to maintain safe operation while still reaching navigation goals. To this end, four experiment setups were considered. 

In the first experiment (Fig. \ref{fig:navigation}a-d), the fire source was turned off. As the fire extinguishing training device does not pose an obstacle to the legged robot, it walked directly over the fire site. This baseline test illustrates that in the absence of heat, the robot is capable of traversing the area without restriction, confirming that the underlying locomotion and perception stack can handle the environment geometrically. In the second experiment (Fig. \ref{fig:navigation}e-h), the fire was ignited and the planner incorporated the thermal radiation field with a large safety margin, causing the robot to take a wide detour around the fire. The third experiment (Fig. \ref{fig:navigation}i-l) repeated this setup but with a smaller safety margin, resulting in a trajectory that approached the heat source more closely while still maintaining safe clearance.

In the fourth experiment (Fig.\ref{fig:wall}a-d), we introduced a wall that occluded the fire. In this case, the planner selected a path adjacent to the wall rather than walking around the fire with a larger distance. This demonstrates that the method prioritizes thermal safety while still optimizing for path efficiency. In Fig. \ref{fig:wall}e-i we show the LiDAR representation along with the planned path. We show the planned path (green) before the fire source was turned on (e) and the new path after the ignition (f). As the wall blocks the fire, our system plans a path through the space adjacent to the wall (g)-(h) to allow the robot to reach its goal (i). In Fig. \ref{fig:wall} we show the frames of the thermal camera along with the measured temperatures. Together, these experiments showcase the adaptability of our proposed framework across varying thermal and geometric constraints.

In Fig. \ref{fig:safety_margin} we show the path planning according to different fire sites and levels of caution. For a small fire in a risk-tolerant setting, the planned navigation trajectory may pass close to the fire (a). This effect can be mitigated by applying a larger caution parameter $\phi$ (b). In contrast, a larger fire results in an expanded danger zone, $q_{\text{danger}}$ (c). With a more conservative choice of $q_{\text{danger}}$, the resulting trajectory takes a longer path around the fire (d). 

A 3D visualization of the estimated thermal radiation profile, computed using our Stefan--Boltzmann model, is shown in Fig.~\ref{fig:temp_elevation}. The figure shows the LiDAR point cloud of the scanned scene in grey, overlaid with color-coded thermal radiation values. The radiation intensity is represented on a continuous scale from blue (lowest) to red (highest), providing a visualization of spatial variations in thermal radiation within the environment.

\begin{figure}[t]
    \centering
    \includegraphics[width=1.0\columnwidth]{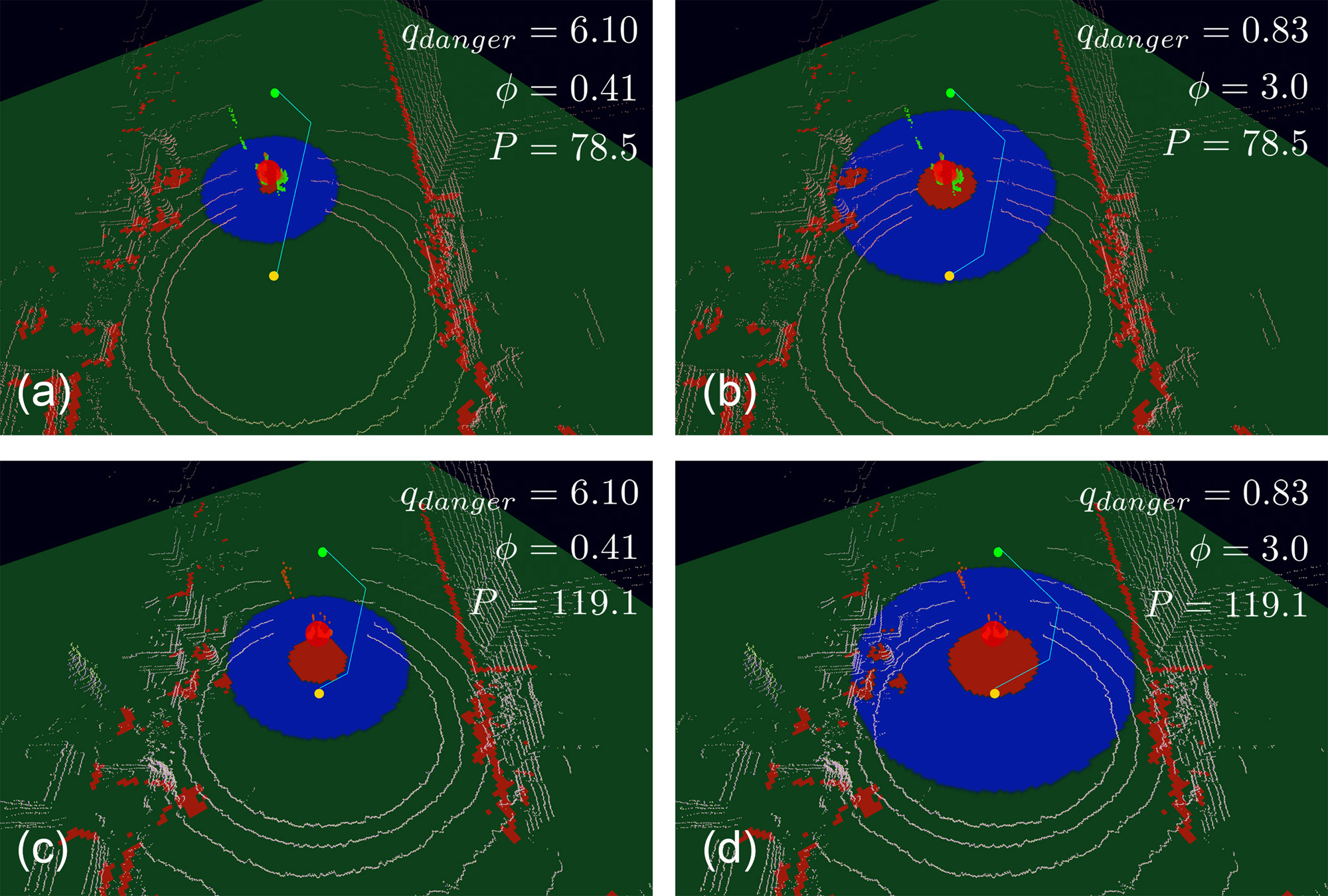}
    \caption{Navigation results for different fire sizes and levels of caution. For a small fire and a risk tolerant setting (a) the planned navigation trajectory closely passes by the fire, which can be mitigated by a larger caution parameter $\phi$ (b). A larger fire with the same $\phi$ as (a) causes a larger danger zone $q_\text{danger}$ (c) which gets expanded even further if we increase $\phi$ (d). Here the blue area indicates cells of the occupancy map with $0.1 < \mathcal{O}[x,y] < 1.0$.}
    \vspace{-4mm}
    \label{fig:safety_margin}
\end{figure}

\subsection{Validation}

To quantify the thermal output of the fire, we conducted a calorimetric experiment using a \(500~\mathrm{ml}\) water sample in a vessel with an exposed surface area of \(0.00975~\mathrm{m^2}\), which we positioned at a distance of \(0.45~\mathrm{m}\) from the fire’s centerline. Over a heating interval of \(722~\mathrm{s}\), the water temperature rose from \(22^{\circ}\mathrm{C}\) to \(42^{\circ}\mathrm{C}\). To ensure uniform heating of the water sample, the vessel was placed on a magnetic stirring plate that continuously mixed the fluid during the experiment.
This measurement yields an absorbed thermal energy of approximately \(4.2\times 10^{4}~\mathrm{J}\). This corresponds to an average net heat transfer rate of \(58~\mathrm{W}\) into the water. To account for environmental losses, we measured the subsequent passive cooling behavior of the water, which exhibited a decrease of \(2.5^{\circ}\mathrm{C}\) every \(600~\mathrm{s}\). This cooling rate corresponds to an ambient heat loss of approximately \(8.7~\mathrm{W}\) at a temperature difference of \(\sim 20~\mathrm{K}\). This implies that average losses during heating were on the order of \(4.4~\mathrm{W}\) (since the water was on average \(\sim 10~\mathrm{K}\) above ambient). 

Adding this to the net absorbed power gives an incident thermal input of about \(62~\mathrm{W}\) on the exposed surface, which corresponds to an incident thermal radiation of \(6.4~\mathrm{kW/m^2}\). Assuming isotropic radiation from the fire, the measured thermal radiation at \(0.45~\mathrm{m}\) implies a total radiative output of approximately 16 kW. Given that open flames, up to roughly \(4~\mathrm{m}\) in diameter,  typically convert about \(30\)--\(40\%\) of their total combustion energy into radiation~\cite{hamins1999nist}, this places the fire’s overall heat release rate in the range of \(38\)–\(53~\mathrm{kW}\).  

We also measured the fuel consumption during the same interval by weighing the gas tank before and after the experiment. The tank mass decreased from \(19.6~\mathrm{kg}\) to \(18.5~\mathrm{kg}\), corresponding to a fuel consumption of \(1.1~\mathrm{kg}\) over the \(722~\mathrm{s}\) test duration. This yields an average mass flow rate of \(\dot{m} = 1.52\times 10^{-3}~\mathrm{kg/s}\). Assuming a typical lower heating value of propane gas of approximately \(46~\mathrm{MJ/kg}\), the corresponding upper limit for the power, assuming a perfect combustion, is \(\dot{Q}_{\mathrm{fuel}} \approx 70~\mathrm{kW}\). 
After accounting for non-perfect combustion, this independent estimate agrees well with the \(38\)--\(53~\mathrm{kW}\) range obtained from the calorimetry analysis and inferred radiative fraction. Our setup is shown in Fig.~\ref{fig:caliometric}. 

When observing the same scenario with our method, we estimate the power of the fire to be 78.5kW (with Eq.~\ref{eq:equation2}, Fig.~\ref{fig:overview}b). Sampling the same 0.45m distance to the fire center yields an estimated thermal radiation of $10.79\mathrm{kW/m^2}$. Using $\phi = 1.0$ and corresponding $q_{danger}=2.5$, this results in a minimum safe distance to the fire of 0.93m while the measurements of $6.4\mathrm{kW/m^2}$ would result in a minimum safe distance of 0.71m. This indicates an error of 0.22m for the estimated safe distance of our model and the measured values. For a detailed review of fluid mechanics of fire we refer the reader to the text of Merci and Beji~\cite{merci2022fluid}.

\section{DISCUSSION AND LIMITATIONS}

Our experiments indicate that explicitly modeling thermal radiation fields from thermal–depth perception enables robots to navigate in fire-affected environments. By localizing the fire source and leveraging a physically inspired thermal radiation decay model, the planner can balance geometric feasibility with thermal safety, producing paths that adapt to varying safety margins and environmental constraints. This  highlights the value of estimating thermal radiation fields and localizing fire sources. A thermal camera by itself is not directly useful for navigation, since it only measures surface radiation at the points visible to the sensor rather than the heat distribution in free space. As a result, the raw thermal image provides no direct estimate of the temperature a robot would experience when moving through the environment. This limitation motivates computing thermal radiation fields, which extends surface measurements into a continuous representation of thermal hazards for safe path planning.

Our approach  provides robots with a representation of fire that can directly be integrated into standard cost-based planning frameworks such as A*.

\begin{figure}[t]
    \centering
    \includegraphics[width=1.0\columnwidth]{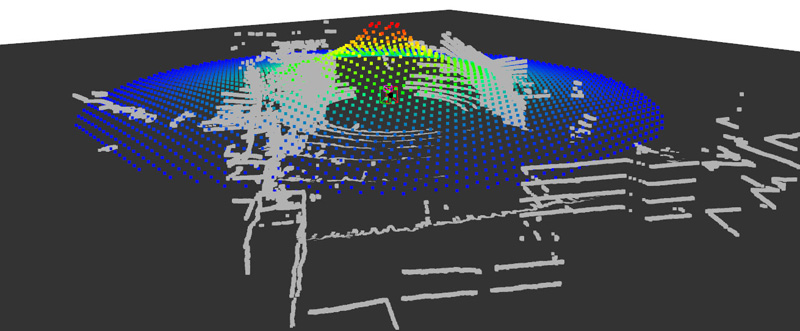}
    \vspace{-4mm}
    \caption{Thermal radiation profile of our model: we show thermal radiation values obtained from a thermal radiation field (colored points) on top of the LiDAR scan of a scene (grey points).}     
    \vspace{-4mm}
    \label{fig:temp_elevation}
\end{figure}
\begin{figure}[t]
    \centering
    \includegraphics[width=1.0\columnwidth]{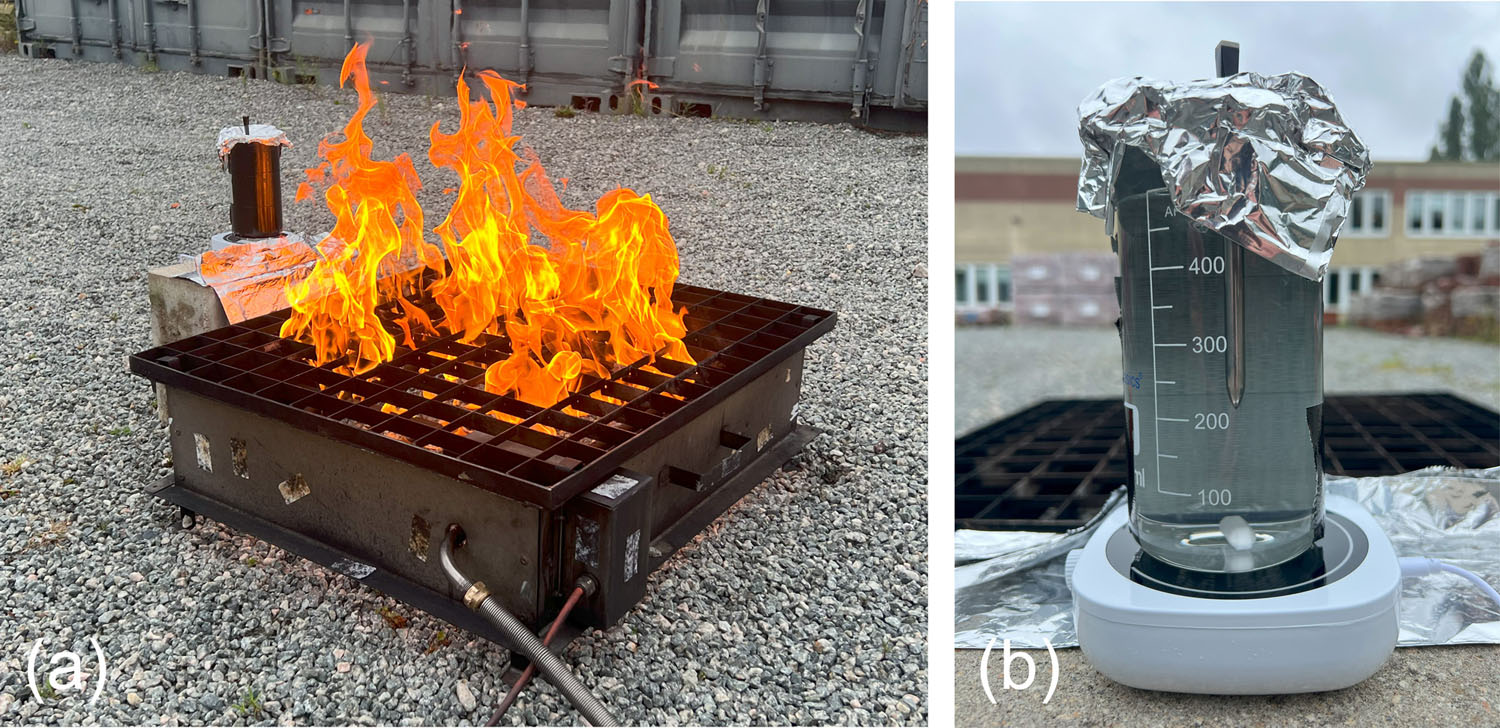}
    \caption{To estimate the thermal radiation, we performed a calorimetry experiment. We measured the time it takes to increase the temperature of 500 ml of water 45 cm away from the center of the fire. }
    \vspace{-5mm}
    \label{fig:caliometric}
\end{figure}

However, our current framework has several limitations. For one, the experiments were performed with a single fire source generated by a propane training device. We have not yet shown that our approach generalizes to multiple or spatially extended fires. Second, the effective strength of the fire was assumed rather than estimated directly, which limits the accuracy of the thermal radiation field under different combustion conditions. Third, our model focuses on radiative heat transfer and does not account for convective effects or more dynamic flame behavior, which can significantly influence heat propagation in real environments. So far, we have only experimented with a common mobile phone attachable thermal camera to obtain radiation measurements. This low-cost camera is only able to measure temperatures up to $192.5 \degree C$. Repeating our experiments with higher-quality thermal sensors would likely lead to better results. We do not assume that stereo reconstructs the transparent flame volume accurately. Rather, we use high-temperature 3D points as a practical proxy for localizing the fire region, including visible flame boundaries and heated surfaces around the burner. While these points contain outliers from mismatched stereo features, a simple clustering algorithm is able to recover accurate flame positions in practice. Robust localization of highly dynamic flames remains a limitation.

\section{CONCLUSIONS}
In this work, we introduced a fire-aware navigation framework that integrates thermal perception into the planning pipeline of a mobile robot. By registering thermal and depth images, we constructed a 3D point cloud annotated with temperature values. We identify the hottest points to localize the fire source, while we employed the Stefan–Boltzmann law to model the spatial decay of heat, yielding a continuous thermal radiation field across the environment. This field was incorporated into a cost map for path planning, allowing the A* algorithm to compute collision-free and thermally safe trajectories. We validated our approach on the Boston Dynamics Spot robot in controlled fire  scenarios with a fire training device, demonstrating that thermal hazards can be explicitly represented and avoided during navigation. The proposed method provides a foundation for deploying autonomous robots in fire-affected environments, with potential applications in firefighting, disaster response, and hazardous inspection tasks. In future work, we plan to extend our modeling to account for dynamic flame behavior, convective heat transfer, and smoke, as well as to test our approach for larger-scale environments and multi-fire scenarios.

%%%%%%%%%%%%%%%%%%%%%%%%%%%%%%%%%%%%%%%%%%%%%%%%%%%%%%%%%%%%%%%%%%%%%%%%%%%%%%%%

%%%%%%%%%%%%%%%%%%%%%%%%%%%%%%%%%%%%%%%%%%%%%%%%%%%%%%%%%%%%%%%%%%%%%%%%%%%%%%%%

%%%%%%%%%%%%%%%%%%%%%%%%%%%%%%%%%%%%%%%%%%%%%%%%%%%%%%%%%%%%%%%%%%%%%%%%%%%%%%%%
\section*{APPENDIX}
All parameters and thresholds (shown in Tab.~\ref{tab:symbols}) are either derived from physical models or obtained from literature.

\section*{ACKNOWLEDGMENTS}
This work has taken place in the VCAI Lab at Kiel University and in the RobotiXX Lab at George Mason University. The work is supported by ERC grant 101170158 - WildfireTwins. RobotiXX research is supported by National Science Foundation (NSF, 2350352), Army Research Office (ARO, W911NF2320004, W911NF2520011), Google DeepMind (GDM), Clearpath Robotics, FrodoBots Lab, Raytheon Technologies (RTX), Tangenta, Mason Innovation Exchange (MIX), and Walmart. We also acknowledge technical support by the Embodied AI Center at Kiel University.

% The preferred spelling of the word ÒacknowledgmentÓ in America is without an ÒeÓ after the ÒgÓ. Avoid the stilted expression, ÒOne of us (R. B. G.) thanks . . .Ó  Instead, try ÒR. B. G. thanksÓ. Put sponsor acknowledgments in the unnumbered footnote on the first page.

\begin{table}[t]
\centering
\vspace{1mm}
\caption{Summary of symbols and parameters used throughout the paper.}
\label{tab:symbols}
\scalebox{0.8}{
\begin{tabular}{p{1.5cm} p{4.2cm} p{2.3cm}}
\toprule
\textbf{Symbol} & \textbf{Description} & \textbf{Units / Notes} \\
\midrule
$r$ & Distance from fire center to query point & m \\

$T_0$ & Effective flame temperature & K \\

$A$ & Estimated emitting surface area (hemisphere approximation) & m$^2$ \\

$\sigma$ & Stefan--Boltzmann constant & $5.670\times10^{-8}$ W\,m$^{-2}$\,K$^{-4}$ \\

$\gamma$ & Empirical correction factor (emissivity + geometry) & 0.4 \\

$P$ & Estimated radiative power of fire source & W \\

$\mathcal{X}$ & Radiative fraction of total heat release & 0.35 \\

$P(r)$ & Radiative heat flux at distance $r$ & W/m$^2$ (or kW/m$^2$) \\

$\mathcal{T}[x,y]$ & Heat-flux value stored in grid cell $(x,y)$ & W/m$^2$ \\

$q_{\text{danger}}$ & Heat-flux danger threshold & W/m$^2$ \\

$\phi$ & Safety margin scaling factor & $>0$ \\

$\mathrm{LoS}(\cdot)$ & Line-of-sight visibility indicator & $\in\{0,1\}$ \\

$\mathcal{O}[x,y]$ & Blended occupancy / hazard value & $[0,1]$ \\

$h_{\text{robot}}$ & Robot height threshold for obstacle filtering & m \\

$C[i,j]$ & Traversal cost for grid cell $(i,j)$ & -- \\

$\beta$ & Cost shaping exponent & $\geq 1$ \\

$\epsilon$ & DBSCAN neighborhood radius & m \\

$m_{\text{min}}$ & DBSCAN minimum samples parameter & integer \\

$K$ & Thermal camera intrinsic matrix & - \\

\bottomrule
\end{tabular}
}
\vspace{-4mm}
\end{table}

%%%%%%%%%%%%%%%%%%%%%%%%%%%%%%%%%%%%%%%%%%%%%%%%%%%%%%%%%%%%%%%%%%%%%%%%%%%%%%%%
\bibliographystyle{IEEEtran}
\bibliography{references}

\end{document}